\documentclass[letterpaper]{article} 
\usepackage{aaai24}  
\usepackage{times}  
\usepackage{helvet}  
\usepackage{courier}  
\usepackage[hyphens]{url}  
\usepackage{graphicx} 
\usepackage{natbib}  
\usepackage{caption} 
\usepackage{algorithm}
\usepackage{algorithmic}
\usepackage{colortbl}
\usepackage{amsmath}
\usepackage{booktabs}
\usepackage{graphicx}
\usepackage{subcaption}
\usepackage{amssymb}
\usepackage{colortbl}
\usepackage[table]{xcolor}

\usepackage{newfloat}
\usepackage{listings}
\DeclareCaptionStyle{ruled}{labelfont=normalfont,labelsep=colon,strut=off} 
\floatstyle{ruled}
\newfloat{listing}{tb}{lst}{}
\floatname{listing}{Listing}
\usepackage[affil-it]{authblk} 

\title{EPNet: An Efficient Pyramid Network for Enhanced Single-Image Super-Resolution with Reduced Computational Requirements}

\author{Xin Xu}
\author{Jinman Park}
\author{Paul Fieguth}
\affil{University of Waterloo}
\date{}
\begin{document}

\maketitle

\begin{abstract}

Single-image super-resolution (SISR) has seen significant advancements through the integration of deep learning. However, the substantial computational and memory requirements of existing methods often limit their practical application. This paper introduces a new Efficient Pyramid Network (EPNet) that harmoniously merges an Edge Split Pyramid Module (ESPM) with a Panoramic Feature Extraction Module (PFEM) to overcome the limitations of existing methods, particularly in terms of computational efficiency. The ESPM applies a pyramid-based channel separation strategy, boosting feature extraction while maintaining computational efficiency. The PFEM, a novel fusion of CNN and Transformer structures, enables the concurrent extraction of local and global features, thereby providing a panoramic view of the image landscape. Our architecture integrates the PFEM in a manner that facilitates the streamlined exchange of feature information and allows for the further refinement of image texture details. Experimental results indicate that our model outperforms existing state-of-the-art methods in image resolution quality, while considerably decreasing computational and memory costs. This research contributes to the ongoing evolution of efficient and practical SISR methodologies, bearing broader implications for the field of computer vision.
\end{abstract}

\section{Introduction}
\label{sec:intro}
Single-Image Super-Resolution (SISR) has become a focal point in the field of computer vision, playing a vital role in a wide variety of applications, including surveillance \cite{zhang2010super, Nascimento_2022}, medical imaging \cite{chen2018efficient, zhu2023residual}, and remote sensing \cite{wang2022comprehensive}. The goal of SISR is to construct a high-resolution image from a single low-resolution counterpart, an inherently ill-posed problem due to the many plausible high-resolution images that can produce the same low-resolution input.

The past several years have seen substantial developments in SISR. Initially, the field was dominated by interpolation methods, such as bicubic \cite{keys} and Lanczos resampling \cite{duchon1979lanczos}. With the progression of research, learning-based methods came into play, utilizing techniques like sparse coding and dictionary learning \cite{Yang}. However, the real transformative shift in the field of SISR came with the emergence of deep learning, and more specifically, convolutional neural networks (CNNs) \cite{lecun1989backpropagation}. CNNs have proven to be highly effective for SISR, learning mappings from low to high-resolution images using training data, and hence, yielding significant improvements in image quality.

\begin{figure}
    \centering
    \includegraphics[width=0.47\textwidth]{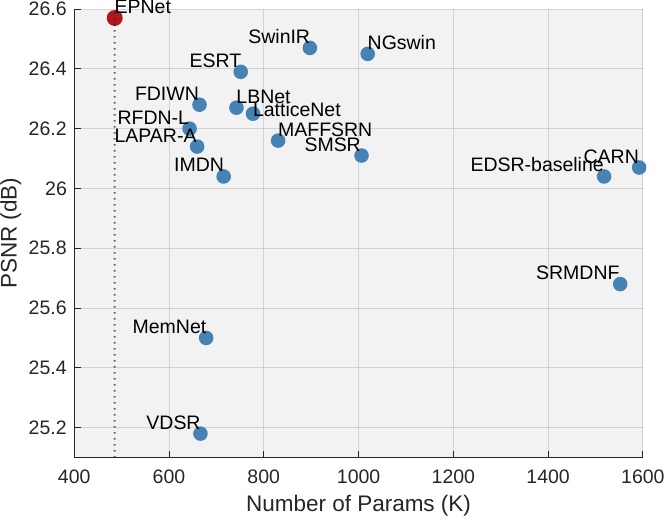}
    \caption{The trade-off between the number of model parameters and performance on Urban100 ($\times4$). Our proposed Efficient Pyramid Network (EPNet) demonstrates exceptional efficiency, achieving a modest parameter count of 485K., while maintaining competitive performance in terms of Peak Signal-to-Noise Ratio (PSNR). The figure highlights the effectiveness of EPNet in striking a balance between model complexity and reconstruction quality, making it a promising solution for SISR tasks.}
    \label{dot_plot}
\end{figure}

Despite these advancements, the real-world application of SISR methods remains challenging due to the high computational and memory costs associated with the complexity of deep learning models. For instance, models like SRCNN \cite{Dong_2014}, EDSR \cite{Lim}, RCAN \cite{Yulun_Zhang}, CARN \cite{Ahn}, IMDN \cite{Hui}, and MADNet \cite{Lan}, while demonstrating impressive performance, suffer from high computational resource consumption. More recently, researchers have been exploring the application of Transformer models, such as the Image Processing Transformer, SwinIR \cite{Liang}, and ESRT \cite{Lu}, to SISR tasks due to their ability to model large receptive field in images. However, these too are constrained by high computational and memory costs.

In response to these challenges, we introduce a lightweight Super-Resolution network tailored for efficiency in SISR tasks. Our network architecture leverages local and global feature extraction through the innovative implementation of two key components: the Edge Split Pyramid Module (ESPM) and the Panoramic Feature Extraction Module (PFEM). In the ESPM, a pyramid-based channel separation strategy is employed for efficient feature extraction \cite{Lin}, including a Dynamic Channel Splitting Attention Block (DCAB) that intelligently reduces the number of feature map channels while preserving useful information, thereby enhancing model efficiency and lightweights. On the other hand, the PFEM forms a composite of CNN and Transformer structures, enabling the concurrent extraction of local and global features. This novel combination provides our network with the capability to effectively navigate and address the complexity of SISR tasks.

\begin{figure*}
    \centering
    \includegraphics[width=\textwidth, trim={0cm 0cm 0cm 0cm},clip]{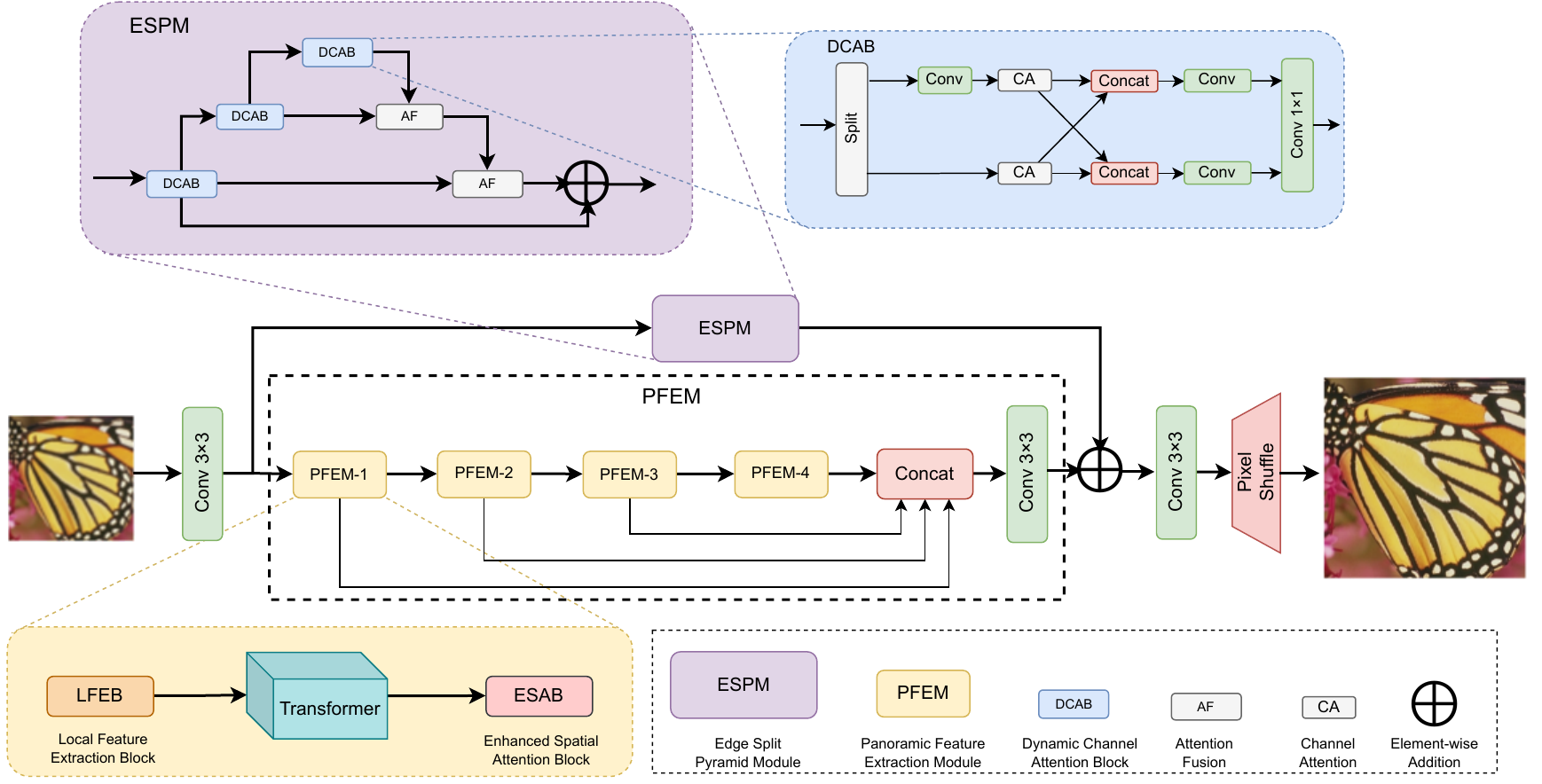}
    \caption{Comprehensive depiction of the proposed Efficient Pyramid Network (EPNet) architecture. This figure is the innovative design and organization of EPNet, which consists of a multi-level pyramid structure, incorporating a judicious combination of Edge Split Pyramid Module (ESPM) and Panoramic Feature Extraction Module (PFEM). This integration enables EPNet to effectively capture and utilize both local and global features, facilitating a panoramic view of the image landscape.}
    \label{architecture}
\end{figure*}

In summary, the main contributions of our paper can be articulated as follows:
\begin{itemize}
\item Introduction of the Edge Split Pyramid Module (ESPM), inspired by the Feature Pyramid Network \cite{lin1}. Tailored for Super-Resolution, ESPM uses a dynamic channel separation strategy to enhance feature extraction while controlling computational and memory costs. This marks a significant advancement in applying Feature Pyramid Network principles to Single-Image Super-Resolution (SISR) tasks.
\item Proposing Panoramic Feature Extraction Module (PFEM), a novel blend of CNNs and Transformer for SISR, complemented by a Local Feature Extraction Block (LFEB) and Enhanced Spatial Attention Block (ESAB) which efficiently captures both local and global image details, revolutionizing the SISR process.
\item Developing Efficient Pyramid Network (EPNet) that integrates ESPM and PFEM, resulting in a high-efficiency SISR architecture. It outperforms existing methods in image quality while reducing computational and memory requirements, paving the way for practical and advanced SISR solutions in computer vision.
\end{itemize}

\section{Related Work}
\label{sec:related}
\subsection{CNN-based SISR Models}

The application of CNN for SISR has seen significant developments. Pioneering work by \citet{Dong} introduced the application of compressed sensing for image super-resolution. Around the same time, \citet{Shi} proposed an efficient sub-pixel convolutional neural network for image super-resolution, which laid a foundation for subsequent research in the field.

Several studies have focused on improving the quality of super-resolved images. \cite{Kim1,Kim2} made considerable progress in accurately restoring high-frequency details in image super-resolution, setting a new standard for the quality of super-resolved images. Later, \citet{Zhang1, Zhang2, Zhang3} proposed image super-resolution using very deep residual channel attention networks, which proved highly effective at leveraging spatial feature information.

Despite these advancements, the challenge of balancing performance and computational cost in CNN-based SISR models remained. Addressing this, \citet{Chen} proposed a method to balance this trade-off, introducing the concept of conditional channel attention for image super-resolution with fewer parameters.

\subsection{Transformer-based SISR Models}

The rise of Transformer models in Natural Language Processing (NLP) has prompted researchers to explore their applications in SISR. Initial efforts faced challenges due to the heavy computational burden of transformers. \citet{Liu} developed a hybrid model that integrated the Transformer with the traditional CNN, overcoming this limitation and highlighting the potential of using Transformers in vision tasks. Furthering this trajectory, \citet{Xiao} addressed the limitations of convolutional operations in contemporary super-resolution (SR) networks by proposing a Self-feature Learning (SFL) mechanism. A lightweight network for image super-resolution adaptively learns image features and effectively mitigates computational costs.

The exploration of Transformer applications in SISR has recently seen significant innovation, particularly with the integration of CNNs and Transformers. A notable development is the Efficient Super-Resolution Transformer (ESRT) \cite{Lu}, which offers a competitive approach to SISR. Following this, \citet{Gao} proposed the Lightweight Bimodal Network (LBNet), a model which adeptly combines a Symmetric CNN for local feature extraction with a Recursive Transformer for learning the long-term dependence of images. The design of LBNet has allowed it to strike an impressive balance between performance, size, execution time, and GPU memory consumption. Recent advancements in SISR, such as ESRT, LBNet, and other cutting-edge methods \cite{zhou2022efficient, li2022blueprint}, have considerably enhanced model performance and efficiency. These methods pose the primary challenge to the approach proposed in this study.

Despite the advancements in both CNN-based and Transformer-based models for SISR, developing models that combine high performance with computational efficiency continues to be a challenge. It is this challenge that our work seeks to address.

\section{Advanced Feature Extraction with Efficient Pyramid Network (EPNet)}
\label{sec:EPNet}
The goal is to formulate a method for efficient and high-quality super-resolution. Given an input low-resolution image $I_{LR}$, the objective is to produce a corresponding high-resolution image $I_{HR}$ that retains the features and quality of a true high-resolution image. In response to this problem, we propose the Efficient Pyramid Network (EPNet), an innovative approach that combines edge detection and panoramic feature extraction to achieve better performance in super-resolution tasks. The rationale for using both edge-specific features and comprehensive features is to create a more robust representation of the image that can enhance the resolution with higher fidelity.

As shown in Figure~\ref{architecture}, the Efficient Pyramid Network (EPNet) is fundamentally composed of three modules, namely, the Edge Split Pyramid Module (ESPM), the Panoramic Feature Extraction Module (PFEM), and an Image Reconstruction mechanism. Notably, ESPM is designed for extracting edge-specific features, while PFEM aims to encapsulate both local and global characteristics of images. We denote $I_{LR}$, $I_{SR}$, and $I_{HR}$ as the input Low-Resolution image, the Super-Resolution reconstructed image, and the corresponding High-Resolution image, respectively.

The model commences its operation by employing a 3x3 convolutional layer for shallow feature extraction:
\begin{equation}
F_{\text{base}} = f_{\text{base}}(I_{LR}),
\end{equation}
where $f_{\text{base}}(\cdot)$ signifies the initial convolutional layer, and $F_{base}$ represents the shallow features extracted. Subsequently, these features are supplied to the PFEM for holistic feature extraction.
\begin{equation}
F_{\text{PFEM}} = f_{\text{PFEM}}(F_{\text{base}}),
\end{equation}
where $f_{\text{PFEM}}(\cdot)$ symbolizes the Panoramic Feature Extraction Module, and $F_{\text{PFEM}}$ represents the comprehensive feature set. The PFEM is a crucial part of the EPNet, comprising multiple pairs of parameter-sharing Local Feature Extraction Blocks (LFEB) and Enhanced Spatial Attention Blocks (ESAB). Detailed descriptions of these components will follow in the succeeding section.

The extracted features are further processed by the ESPM for edge feature learning:
\begin{equation}
F_{\text{ESPM}} = f_{\text{ESPM}}(F_{\text{base}}),
\end{equation}
where $F_{\text{ESPM}}$ denotes the edge detail feature extracted by the ESPM $f_{\text{ESPM}}(\cdot)$. 

Finally, the refined features from $F_{\text{ESPM}}$ and the shallow features from $F_{\text{PFEM}}$ are combined and supplied to the reconstruction module to yield the Super Resolution (SR) image:
\begin{equation}
I_{SR} = Rec_{\text{module}}(F_{\text{PFEM}} + F_{\text{ESPM}}),
\end{equation}
where $Rec_{\text{module}}(\cdot)$ denotes the image reconstruction module, which comprises a 3x3 convolutional layer and a pixel-shuffle layer.

EPNet is trained with an L1 loss optimization strategy. Given a training dataset $\{I_{LR}^{i}, I_{HR}^{i}\}_{i=1}^{N}$, we aim to minimize the following loss function:
\begin{equation}
\hat{\theta} = \arg\min_{\theta} \frac{1}{N} \sum_{i=1}^{N} \| Rec_{\theta}(I_{LR}^{i}) - I_{HR}^{i} \|_1,
\end{equation}
where $\theta$ represents the set of parameters associated with our proposed EPNet, $Rec_{\theta}(I_{LR}) = I_{SR}$ corresponds to the SR image reconstruction, and $N$ is the number of images in the training dataset.

\begin{figure}
    \centering
    \includegraphics[width=0.44\textwidth]{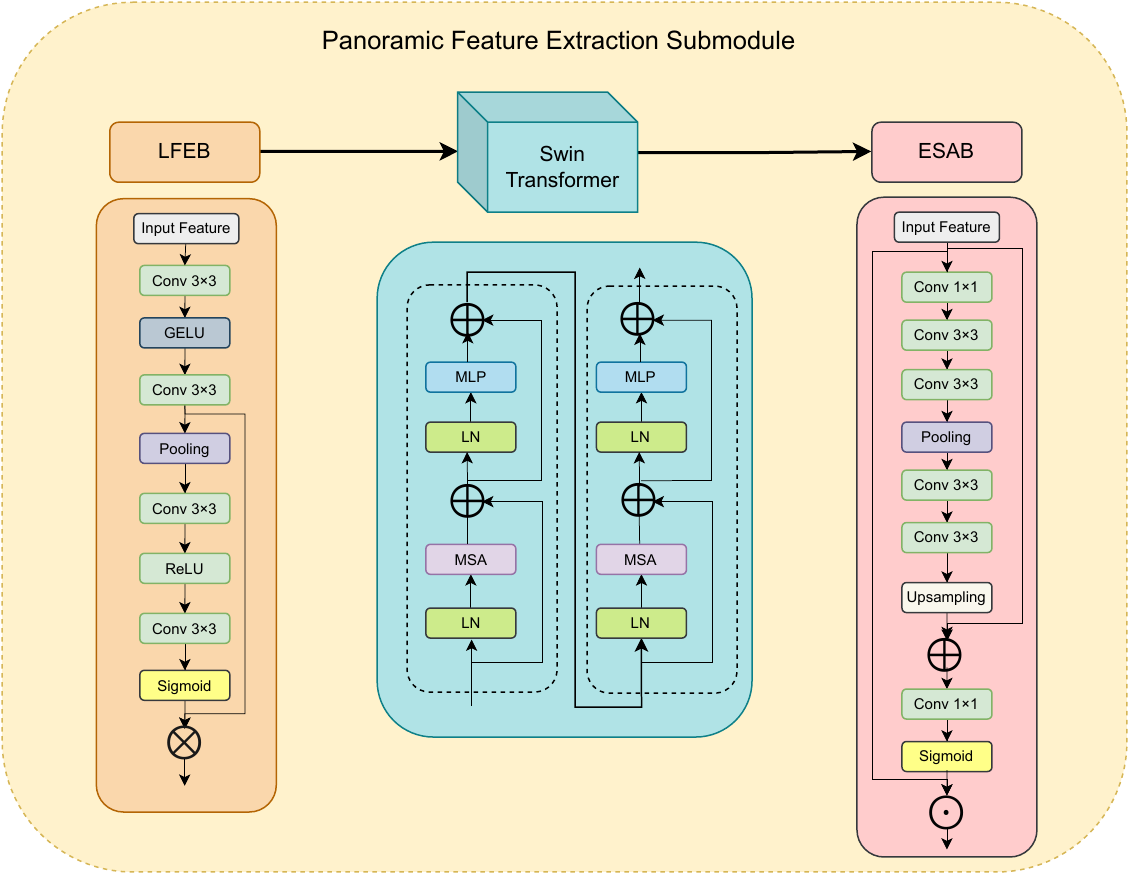}
    \caption{Panoramic Feature Extraction Module (PFEM) implementation. PFEM comprises LFEB, Swin Transformer, and ESAB, enabling simultaneous extraction of local and global features.}
    \label{detailed}
\end{figure}

\subsection{Edge Split Pyramid Module (ESPM)}

The Edge Split Pyramid Module (ESPM), a central part of our proposed architecture, builds upon the principles of the Feature Pyramid Network \cite{lin1}, with particular adjustments to address the complexities of Single-Image Super-Resolution (SISR) tasks.

\subsubsection{Dynamic Channel Attention Block (DCAB)}

The Dynamic Channel Attention Block (DCAB) within the ESPM is designed to enhance efficiency and effectiveness by balancing the extraction of semantic information with computational constraints. This is achieved through a process known as dynamic channel separation, allowing for more nuanced handling of different features within the image.

The DCAB seeks to emphasize vital information while minimizing computational and memory overhead. By selectively reducing the number of feature map channels and retaining only essential information, the DCAB strikes a balance between efficiency and performance.

\paragraph{Weighted Combinatorial Crossover}

A novel aspect of the DCAB's operation is the use of a weighted combinatorial crossover. This technique can be expressed as:
\begin{equation}
F_{\text{DCAB}} = \text{concat} \left( \text{concat}\left( \alpha \cdot x_1, x_2 \right), \text{concat}\left( x_1, \beta \cdot x_2 \right) \right),
\end{equation}
where the input feature map $x$ is bifurcated into parts $x_1$ and $x_2$, and modulated by channel attentions $\alpha$ and $\beta$, respectively. This allows for targeted processing of different features, enabling specific treatments based on their relevance to the reconstruction task. The channel attentions $\alpha$ and $\beta$ serve to weight the bifurcated parts, underscoring essential features and suppressing others. 

\subsection{Panoramic Feature Extraction Module (PFEM)}

The Panoramic Feature Extraction Module (PFEM) is designed to surmount the inherent limitations of CNNs in understanding the global context within images. While CNNs excel in extracting local features, they tend to overlook broader relationships across the image, a challenge that becomes crucial in tasks such as SISR. The PFEM addresses this by incorporating CNNs for local feature extraction, recognizing their efficiency in this specific aspect, but then extends their capability through integration with Transformers.

In this combined architecture, Transformers play a vital role in building a global perspective that encompasses the entire image. Unlike CNNs, Transformers can evaluate the relationships between all parts of the image, linking the fine-grained local features extracted by CNN layers and placing them within the global context. This synthesis within the PFEM enables the Efficient Pyramid Network (EPNet) to provide detailed local features enriched with an understanding of the broader image structure. By doing so, the PFEM not only mitigates the constraints of CNNs but amplifies their strengths, providing the model with a more comprehensive and nuanced view of the input image's characteristics, leading to enhanced performance in SISR tasks.

The PFEM is composed of n submodules. The output of the module is calculated as follows:
\begin{equation}
    F_\text{PFEM} = \phi^n(\phi^{n-1}(...(\phi^1(F_\text{base}))))
\end{equation}
where $\phi_i(\cdot)$ represents the operation of the i-th submodule and $F_\text{PFEM}$ represents the feature map obtained after the n-th submodule operation. The optimal balance of performance and consumption is achieved when $n=4$, which we utilize in our model. Detailed discussion and analysis related to this selection will be provided in the experimental section.

As shown in Figure~\ref{detailed}, the PFEM consists of three primary components: the Local Feature Extraction Block (LFEB), a slightly modified variant of the Swin Transformer \cite{Swin}, and the Enhanced Spatial Attention Block (ESAB). The LFEB leverages the power of CNNs to extract local features, while the modified Swin Transformer adapts the Transformer's ability to capture global context. Lastly, the ESAB facilitates the EPNet's focus on the most salient areas of the image. Through this carefully orchestrated module, the PFEM ensures comprehensive and effective feature extraction, making the EPNet a potent tool for SISR tasks.

\subsubsection{Local Feature Extraction Block (LFEB)}

A fundamental component of the Panoramic Feature Extraction Module (PFEM) is the Local Feature Extraction Block (LFEB). This block is structured around a convolutional network, with an emphasis on spatial feature interactions.

The distinctive feature of LFEB is the implementation of an Enhanced Channel Attention Module (ECAM), inspired by the Efficient Channel Attention strategy in ECA-Net \cite{wang2020eca}. Unlike standard global average pooling methods, the ECAM operates without causing dimensionality reduction, allowing it to maintain more information from the input. This enables the ECAM to grasp the local cross-channel dependencies, where each channel is analyzed in the context of its neighboring channels, thereby capturing the relationship and significance among them.

Such design targets the effective modulation of the attention map, selectively emphasizing features that are more relevant to the task while reducing the influence of less pertinent ones. This alignment between channel relationships and task-specific importance shapes LFEB as a competent tool for detailed local feature extraction. It functions not by the sheer complexity but by aligning the extraction process with the inherent properties of SISR, thus contributing substantially to the overall performance of the PFEM, and by extension, the EPNet.

\subsubsection{Enhanced Spatial Attention Block (ESAB)}
The Enhanced Spatial Attention Block (ESAB), integrated into the Panoramic Feature Extraction Module (PFEM), draws from the Convolutional Block Attention Module (CBAM) \cite{Woo} and Residual Networks (ResNets) \cite{he2016deep}. This component of our model is meticulously engineered to harness spatial attention mechanisms, enabling the extraction of expansive global information. The rationale for this design lies in its ability to complement the Local Feature Extraction Block (LFEB)'s concentration on intricate local details, creating a harmonious representation of both global and local features. The collaboration of these elements leads to a more comprehensive and accurate portrayal of the underlying image data.

Within the ESAB, the architecture is initiated with a convolution operation followed by a sequence of convolutional layers that progressively refine the processed global features. This step-by-step refinement represents a carefully considered strategy aimed at preserving the broader context of the image while allowing for precise adjustments. Additionally, the incorporation of a max pooling operation effectively isolates and emphasizes the most salient global features, an approach that is instrumental in sharpening the global feature representation. By aligning these elements, the ESAB's design illustrates an intelligent and deliberate response to the specific challenges of single-image super-resolution tasks, demonstrating how each component serves a clear purpose in achieving a balanced and high-quality output.

\section{Experiments}
\label{sec:experiments}

\begin{table}[b]
\centering
\small
\begin{tabular}{cc|cccc}
\toprule 
Method & Year & Params & Multi-Adds & PSNR/SSIM \\
\midrule
SwinIR & 2021 & 897K & 49.6G & \textbf{29.26}/\textbf{0.8274}  \\
ESRT & 2022 &  751K & 67.7G & 29.14/0.8244 \\
LBNet & 2022 & 742K & 38.9G & 29.12/0.8238 \\
NGswin & 2023 & 1019K & 36.4G & 29.20/0.8262 \\
\rowcolor[gray]{0.9} \textbf{EPNet (ours)} & 2023 & \textbf{485K} & \textbf{23.3G} & \textbf{29.26}/0.8271 \\

\bottomrule
\end{tabular}
\caption{Quantitative comparison of SOTA LSR methods on benchmark datasets of $\times4$. The table showcases the average PSNR and SSIM across five datasets. Best performing metrics are \textbf{highlighted}.}
\label{multi-adds}
\end{table}

\subsection{Datasets and Metrics}
In our study, we utilized a range of diverse datasets and evaluation metrics to ensure a comprehensive analysis of our model's performance. Our model was trained on the DIV2k dataset \cite{Timofte}, a standard benchmark for super-resolution that includes one thousand 2K resolution RGB images. The diversity and high quality of the DIV2k dataset provided a rich foundation for our model's learning. To validate the model's effectiveness, we employed several testing sets, including Set5 \cite{Bevilacqua}, Set14 \cite{Yang}, BSDS100 \cite{Martin}, Urban100 \cite{Huang}, and Manga109 \cite{Aizawa}, each presenting unique challenges and content types. To measure the performance of our model, we used the Peak Signal-to-Noise Ratio (PSNR) and Structural Similarity Index Measure (SSIM) \cite{zhou_wang}, two widely accepted metrics in the field of Single-Image Super-Resolution (SISR). PSNR quantifies the amount of reconstruction error, with higher values indicating superior image quality, while SSIM provides a comprehensive evaluation of image quality based on changes in structural information, luminance, and contrast.

\begin{table*}[ht]
\centering
\small
\begin{tabular}{lccccccc}
\hline
\\[-2ex]
Method  & Params & Set5 & Set14 & BSD100 & Urban100 & Manga109 \\
\\[-2ex]
\hline
\\[-2ex]
Scale {$\times3$} & & {PSNR/SSIM} & {PSNR/SSIM} & {PSNR/SSIM} & {PSNR/SSIM} & {PSNR/SSIM} \\
\\[-2ex]
\hline
\\[-2ex]
VDSR \cite{Kim1} & 666K & 33.66/0.9213 & 29.77/0.8314 & 28.82/0.7976 & 27.14/0.8279 & 32.01/0.9340 \\
MemNet \cite{Tai}  & 678K & 34.09/0.9248 & 30.00/0.8350 & 28.96/0.8001 & 27.56/0.8376 & 32.51/0.9369 \\
EDSR-baseline \cite{Lim}  & 1,555K & 34.37/0.9270 & 30.28/0.8417 & 29.09/0.8052 & 28.15/0.8527 & 33.45/0.9439 \\
SRMDNF \cite{Kai_Zhang}  & 1,528K & 34.12/0.9254 & 30.04/0.8382 & 28.97/0.8025 & 27.57/0.8398 & 33.00/0.9403 \\
CARN \cite{Ahn}  & 1,592K & 34.29/0.9255 & 30.29/0.8407 & 29.06/0.8034 & 28.06/0.8493 & 33.50/0.9440 \\
IMDN \cite{Hui}  & 703K & 34.36/0.9270 & 30.32/0.8417 & 29.09/0.8046 & 28.17/0.8519 & 33.61/0.9445 \\
RFDN-L \cite{Liu}  & 633K & 34.47/0.9280 & 30.35/0.8421 & 29.11/0.8053 & 28.32/0.8547 & 33.78/0.9458 \\
MAFFSRN \cite{Muqeet}  & 807K & 34.45/0.9277 & 30.40/0.8432 & 29.13/0.8061 & 28.26/0.8552 & -/- \\
LatticeNet \cite{Luo}  & 765K & \textbf{34.53}/\underline{0.9281} & 30.39/0.8424 & \underline{29.15}/0.8059 & 28.33/0.8538 & -/- \\
SMSR \cite{Wang}  & 993K & 34.40/0.9270 & 30.33/0.8412 & 29.10/0.8050 & 28.25/0.8536 & 33.68/0.9445 \\
LAPAR-A \cite{li}  & 594K & 34.36/0.9267 &  30.34/0.8421 & 29.11/0.8054 & 28.15/0.8523 &  33.51/0.9441 \\
ESRT \cite{Lu}  & 770K & 34.42/0.9268 & \underline{30.43}/\underline{0.8433} & \underline{29.15}/\underline{0.8063} & \underline{28.46}/\underline{0.8574} & \underline{33.95}/0.9455 \\
LBNet \cite{Gao}  & 736K & 34.47/0.9277 & 30.38/0.8417 & 29.13/0.8061 & 28.42/0.8559 & 33.82/\underline{0.9460} \\
FDIWN \cite{gao2} & 645K & 34.52/0.9281 & 30.42/0.8438 & 29.14/0.8065 & 28.36/0.8567 & -/-\\
\rowcolor[gray]{0.9} \textbf{EPNet(ours)}  & \textbf{475K} & \underline{34.49}/\textbf{0.9284} & \textbf{30.46}/\textbf{0.8448} & \textbf{29.16}/\textbf{0.8082} & \textbf{28.50}/\textbf{0.8590} & \textbf{33.96}/\textbf{0.9466} \\

\\[-2ex]
\hline
\\[-2ex]
Scale {$\times4$} &  & {PSNR/SSIM} & {PSNR/SSIM} & {PSNR/SSIM} & {PSNR/SSIM} & {PSNR/SSIM} \\
\\[-2ex]
\hline
\\[-2ex]
VDSR \cite{Kim1}  & 666K & 31.35/0.8838 & 28.01/0.7674 & 27.29/0.7251 & 25.18/0.7524 & 28.83/0.8870 \\
MemNet \cite{Tai} & 678K & 31.74/0.8893 & 28.26/0.7723 & 27.40/0.7281 & 25.50/0.7630 & 29.42/0.8942 \\
EDSR-baseline \cite{Lim}  & 1,518K & 32.09/0.8938 & 28.58/0.7813 & 27.57/0.7357 & 26.04/0.7849 & 30.35/0.9067 \\
SRMDNF \cite{Kai_Zhang}  & 1,552K & 31.96/0.8925 & 28.35/0.7787 & 27.49/0.7337 & 25.68/0.7731 & 30.09/0.9024 \\
CARN \cite{Ahn}  & 1,592K & 32.13/0.8937 & 28.60/0.7806 & 27.58/0.7349 & 26.07/0.7837 & 30.47/0.9084 \\
IMDN \cite{Hui}  & 715K & 32.21/0.8948 & 28.58/0.7811 & 27.56/0.7353 & 26.04/0.7838 & 30.45/0.9075 \\
RFDN-L \cite{Liu}  & 643K & 32.28/0.8957 & 28.61/0.7818 & 27.58/0.7363 & 26.20/0.7883 & 30.61/0.9096 \\
MAFFSRN \cite{Muqeet}  & 830K & 32.20/0.8953 & 28.62/0.7822 & 27.59/0.7370 & 26.16/0.7887 & -/- \\
LatticeNet \cite{Luo}  & 777K & \textbf{32.30}/\underline{0.8962} & 28.68/0.7830 & 27.62/0.7367 & 26.25/0.7873 & -/- \\
SMSR \cite{Wang}  & 1006K & 32.12/0.8932 & 28.55/0.7808 & 27.55/0.7351 & 26.11/0.7868 &  30.54/0.9085 \\
LAPAR-A \cite{li}  & 659K & 32.15/0.8944 & 28.61/0.7818 & 27.61/0.7366 & 26.14/0.7871 & 30.42/0.9074 \\
ESRT \cite{Lu}  & 751K & 32.19/0.8947 & \underline{28.69}/\underline{0.7833} & \textbf{27.69}/0.7379 & \underline{26.39}/\underline{0.7962} & 30.75/0.9100 \\

LBNet \cite{Gao}  & 742K & \underline{32.29}/0.8960 & 28.68/0.7832 & 27.62/\underline{0.7382} & 26.27/0.7906 & \underline{30.76}/\underline{0.9111} \\
FDIWN \cite{gao2} & 664K & 32.23/0.8955 & 28.66/0.7829 & 27.62/0.7380 & 26.28/0.7919 & -/-\\

\rowcolor[gray]{0.9} \textbf{EPNet(ours)}  & \textbf{485K} & 32.28/\textbf{0.8964} & \textbf{28.74}/\textbf{0.7849} & \underline{27.67}/\textbf{0.7404} & \textbf{26.57}/\textbf{0.7995} & \textbf{31.02}/\textbf{0.9145} \\
\hline
\end{tabular}
\caption{Comparison of Methods for Super Resolution at Scales $\times$3 and $\times$4, where the scale value indicates the factor by which the low-resolution image is upscaled. The best and second best results are \textbf{highlighted} and \underline{underlined}, respectively.}
\label{table2}

\end{table*}

\subsection{Comparisons with Lightweight SISR Models}

\begin{figure*}
    \centering
    \includegraphics[width=\textwidth, trim={0cm 1cm 0cm 0cm},clip]{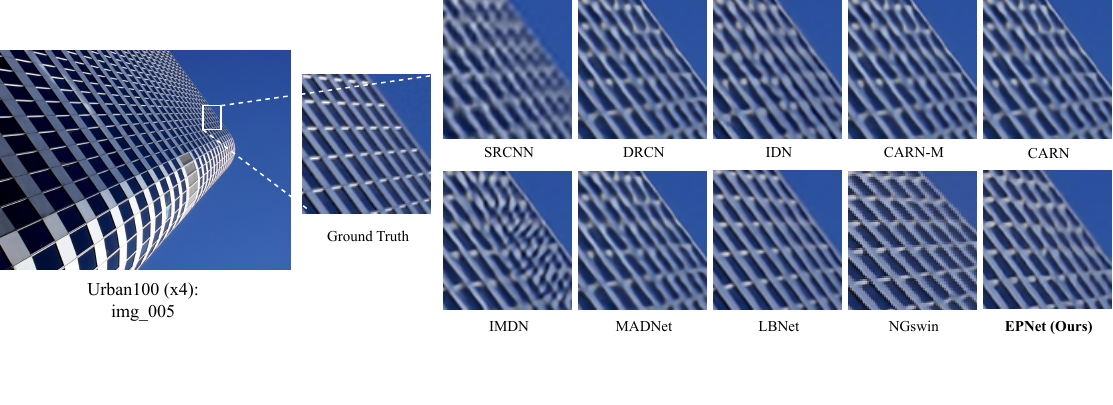}
    \caption{Visual Comparison with Lightweight SISR Models. EPNet outperforms other methods, offering superior image resolution with low parameters. EPNet strikes a balance between complexity and performance, making it an efficient choice for resource-constrained applications.}
    \label{visual}
\end{figure*}

In Table~\ref{table2}, we present a comprehensive comparison between our model, EPNet, and various other state-of-the-art lightweight SISR models. It is quite evident that our model demonstrates exceptional performance, achieving optimal or near-optimal results across various scales and image datasets, while maintaining a considerably smaller model size. Remarkably, EPNet achieves a strong balance between model performance and model size, sporting only 475K and 485K parameters for ×3 and ×4 scales, respectively.

Further accentuating our model's robustness is the visual comparison between EPNet and other lightweight SISR models (as shown in Figure~\ref{visual}). Clearly, EPNet displays richer, more detailed textures, yielding superior visual effects. These observations serve as strong validation for the effectiveness of our proposed EPNet model.

The performance of EPNet is juxtaposed with other models in Table~\ref{multi-adds}, considering factors such as PSNR, parameter quantity, and Multi-Add operations. Despite securing competitive PSNR results, EPNet distinguishes itself by maintaining relatively low values for Multi-Adds and the parameter count. This evidence consolidates EPNet's standing as an efficient, lightweight SISR model.

\subsection{Model Complexity Studies}

When assessing the performance of Learned Single Image Super-Resolution (LSISR) models, it's crucial to consider several facets, not just the PSNR and SSIM scores. Model size and computational efficiency also play pivotal roles, particularly when contemplating the practical deployment of these models. Striking the right balance between performance and computational requirements is key.

In this context, our proposed model, EPNet, excels, as illustrated by the comparison table provided. The table allows for an easy comparison of several state-of-the-art models, highlighting how each fares in terms of complexity and performance. In Table 2, EPNet stands out for its ability to deliver exceptional results without excessive computational costs. With a moderate model size of 485K parameters and 23.3G Multi-Adds, EPNet achieves a remarkable PSNR/SSIM score of 29.26 / 0.827. This performance is significant considering that other models like SwinIR \cite{Liang}, ESRT \cite{Lu}, LBNet \cite{Gao}, and NGswin \cite{choi2023n} necessitate substantially greater computational resources for achieving similar or even lower performance levels.

Interestingly, despite leveraging a Transformer architecture, which is generally known for its higher computational demands, EPNet manages to maintain an impressive level of efficiency. What sets EPNet apart is its hybrid design that goes beyond the conventional Transformer usage seen in other models. It ingeniously combines the strengths of the Swin Transformer \cite{Swin} with innovative channel-splitting techniques in the Edge split Pyramid Module (ESPM). This novel approach helps reduce the model size without compromising performance, thereby achieving similar or even superior results compared to other methods.

When we compare EPNet with models such as LBNet and NGswin, EPNet not only provides comparable or better performance but does so with a significantly smaller set of parameters. This is indicative of EPNet's superior efficiency-performance trade-off, as it delivers high-quality super-resolution images without corresponding increases in computational resources. This characteristic makes EPNet an appealing choice for lightweight SISR tasks.

\subsection{Ablation Study}

\begin{figure}[H]
    \centering
    \begin{subfigure}{0.22\textwidth}
        \centering
        \includegraphics[width=\textwidth]{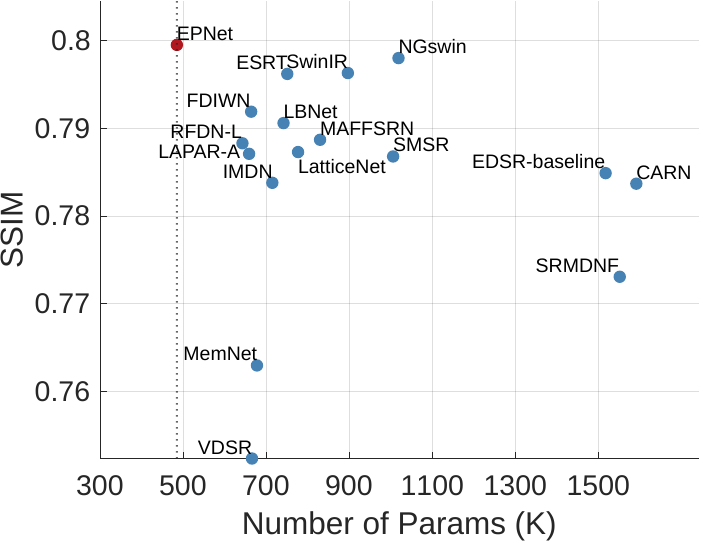}
        \caption{Caption for the first subfigure}
        \label{complexity1}
    \end{subfigure}
    \hfill
    \begin{subfigure}{0.22\textwidth}
        \centering
        \includegraphics[width=\textwidth]{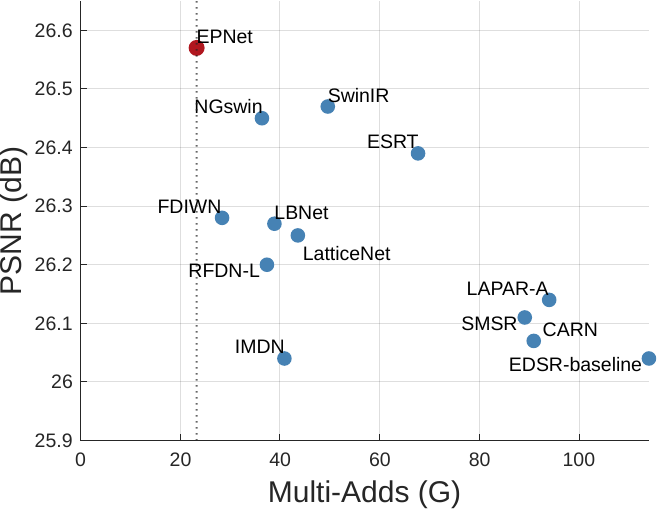}
        \caption{Caption for the second subfigure}
        \label{complexity2}
    \end{subfigure}
    \caption{Model complexity study on Urban100 at a ($\times4$) scale. With 485K parameters and 23.3G multi-adds, EPNet achieves a PSNR of 29.26 and SSIM of 0.8271. The figure illustrates EPNet's balanced architecture, achieving competitive results while maintaining low complexity in comparison with other methods.}
    \label{complexity}
\end{figure}

\begin{table}[H]
\centering
\small
\begin{tabular}{cccccc}
\toprule 
Scale & ESPM & ESAB & LFEB & Params & PSNR/SSIM \\
\midrule

$\times$2 & $\checkmark$ & $\checkmark$ & $\times$ & 318K & 32.11/0.9290 \\

$\times$2 & $\checkmark$ & $\times$ & $\checkmark$ & 368K & 32.05/0.9240 \\

$\times$2 & $\times$ & $\checkmark$ & $\checkmark$ & 388K & 31.75/0.9251 \\

\rowcolor[gray]{0.9} $\times$2 & $\checkmark$ & $\checkmark$ & $\checkmark$ & 468K & \textbf{32.29/0.9299} \\
\bottomrule
\end{tabular}
\caption{Ablation Study of EPNet Components on Urban100 ($\times$2). The study highlights the crucial role of ESPM, ESAB, LFEB in achieving superior image enhancement, underscoring the effectiveness of EPNet's architecture for efficient and high-quality SISR.}
\label{ablation}
\end{table}

\subsubsection{Evaluation of Each Component in EPNet}
To gauge the performance and contribution of each component within our EPNet model, we undertook an ablation study on the Urban100 dataset at a scale of ×2. The choice of scale ×2 was motivated by its representativeness of typical use cases in SISR tasks. It provides a balanced challenge, being neither too simplistic nor overly complex, thus offering a fair evaluation of the model's capabilities. The results of this study are detailed in Table~\ref{ablation}.

The findings reveal that each component (ESPM, ESAB, and LFEB) plays a significant role in the performance of the model. The best PSNR and SSIM metrics are obtained when all components are included in the model.

\subsubsection{Panoramic Feature Extraction Module Investigations}
We further investigated the Panoramic Feature Extraction Module (PFEM) within the EPNet model by conducting experiments with varying numbers of PFEMs. Table 3 presents the results of this investigation.

The results demonstrate that with an increase in the number of PFEMs, both the PSNR and SSIM values improve. This indicates the significant role of PFEMs in enhancing the model's performance on both Set5 and Manga109 datasets. However, the increase in the number of PFEMs also leads to a rise in the number of parameters, implying an increase in computational cost.

In light of these results, we have chosen to set the number of PFEM blocks to $n=4$ in our final model. This decision strikes a balance between computational cost and model performance, taking into consideration the performance gains with each additional PFEM against the corresponding increase in computational load.

Through these ablation studies, we have validated the effectiveness of each component in the EPNet architecture, and the role of the Panoramic Feature Extraction Module in particular. These findings emphasize the robustness and efficiency of our proposed model.

\section{Conclusion}
This paper presented EPNet, a novel and efficient architecture for Single-Image Super-Resolution (SISR), integrating innovative components - the Edge Split Pyramid Module (ESPM) and the Panoramic Feature Extraction Module (PFEM). The ESPM introduces a pyramid-based channel separation strategy, enhancing feature extraction efficiency, while the PFEM combines the strengths of Convolutional Neural Networks and Transformer structures, efficiently capturing both local and global image features. Empirical assessments validate EPNet's superior performance against state-of-the-art methods, accomplishing competitive image resolution quality with significantly reduced computational and memory demands. This breakthrough underscores the potential for practical, efficient, and high-performance SISR solutions, with wider implications for various computer vision tasks demanding efficient high-resolution imaging. Future endeavors may explore potential improvements in component design and broader application contexts.

\bibliography{aaai24}
\newpage

\appendix
\section{Appendix}

\subsection{Implementation Details}
The Efficient Pyramid Network (EPNet) is developed using the PyTorch framework and trained on a standalone GTX 3090Ti GPU. Our training protocol involves extracting image patches of size $48\times48$ and training with a batch size of 32. In the image reconstruction component of EPNet, we adhere to prevalent practices, employing PixelShuffle to enhance the last coarse features to finer features.

During the training process, the Exponential Moving Average (EMA) technique with a decay rate of 0.999 is utilized, which provides increased stability to the model. The EMA is defined as:
\begin{equation}
    x_n = \theta_1 - \sum_{i=1}^{n-1} (1-\beta^{n-i}) g_i
\end{equation}
Where the \(x_n\) is the updated parameter at the n-th step and $\theta_1$ is the initial parameter value before any updates have been made. The hyperparameter \(\beta\) is the learning decay rate. The term \(g_i\) represents the gradient at the i-th step.

The model is optimized with the Adam optimizer \cite{adam}, configured with a learning rate of $5\times10^{-4}$, $\beta$ set to 0.9 and 0.99, and devoid of weight decay. We train the model over a total of $1\times10^{6}$ iterations without a warm-up phase. Furthermore, L1 loss is leveraged during training as it facilitates the generation of sharper images.

\subsection{Description of the Figure}
\begin{figure}[h]
    \centering

    \begin{subfigure}{0.21\textwidth}
        \includegraphics[width=\textwidth]{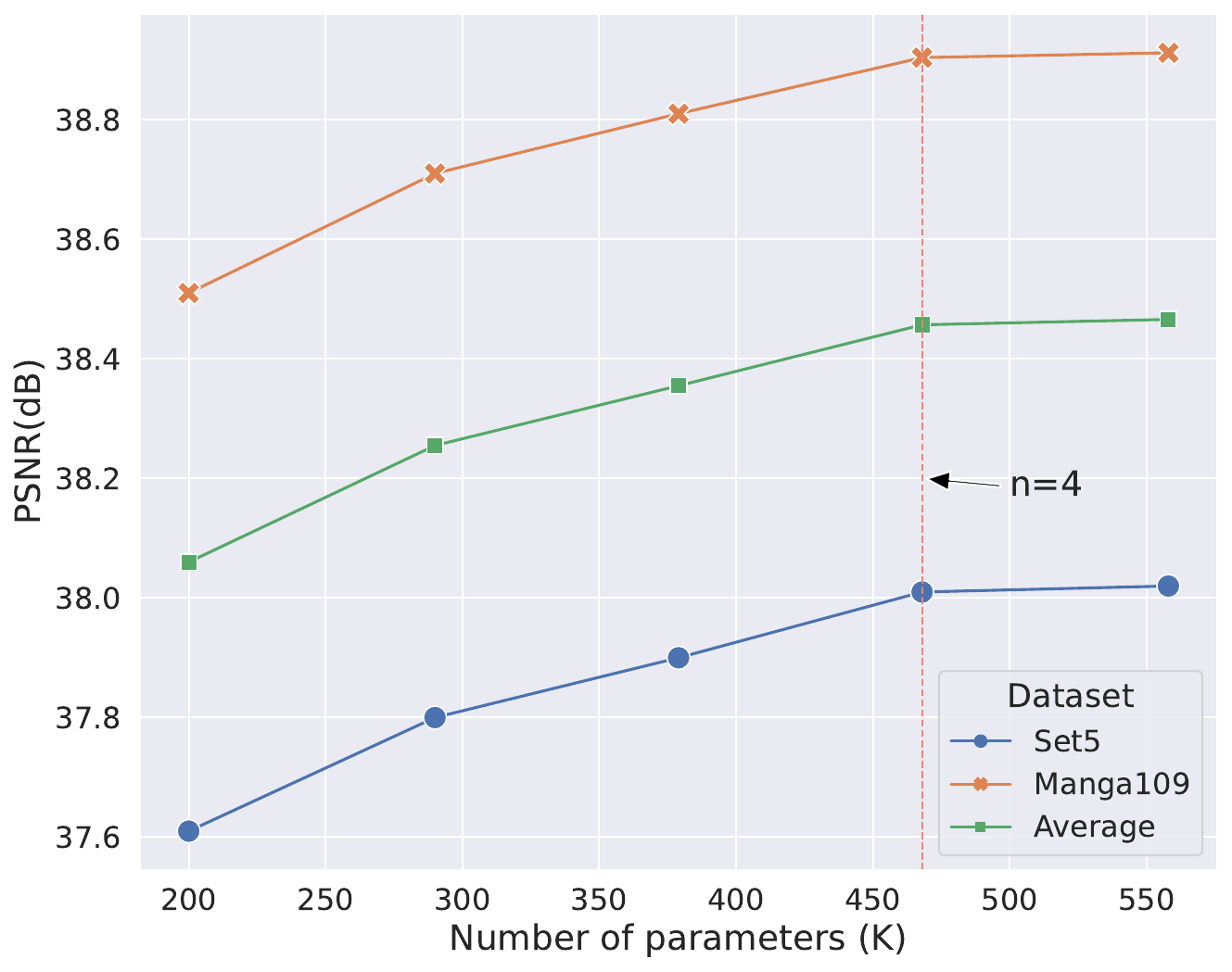}
        \label{PFEM1}
    \end{subfigure}
    \hfill
    \begin{subfigure}{0.238\textwidth}
        \includegraphics[width=\textwidth]
        {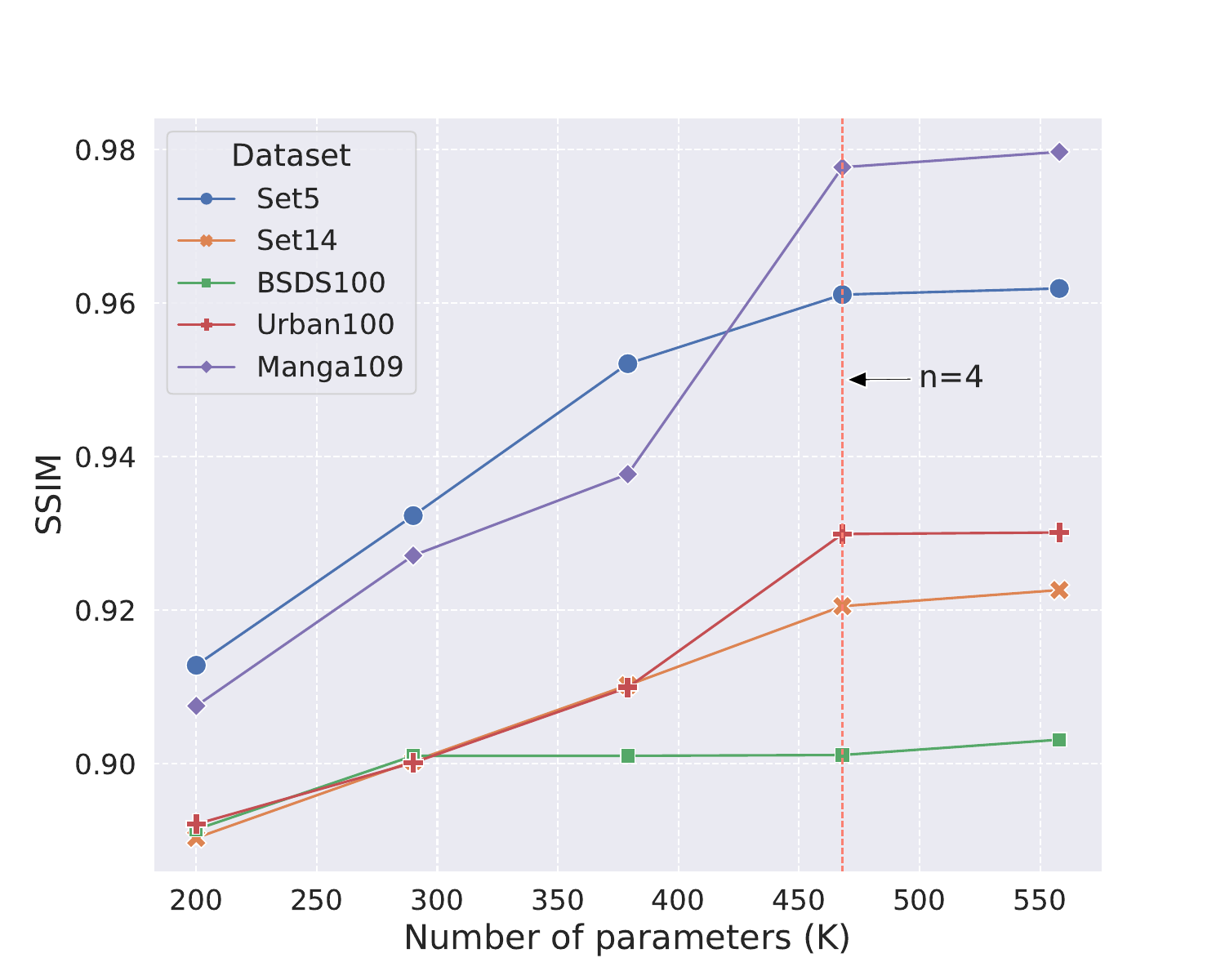}
        \label{PFEM2}
    \end{subfigure}
    \caption{Study of the number of PFEM with the complexity and performance on Set5 and Manga109. Notably, with $n=4$ PFEM modules, the EPNet achieves the best trade-off between model complexity (parameters) and PSNR performance. Beyond this point, further increases in PFEM modules yield diminishing returns in performance improvement.}
    \label{PFEM}

\end{figure}

Figure \ref{PFEM} showcases a study of the number of PFEM modules in relation to the complexity and performance metrics. Specifically, we can observe that as the number of PFEM modules reaches 4, the EPNet exhibits an optimal balance between model complexity and PSNR performance. Incorporating more than 4 PFEM modules doesn't seem to offer significant performance enhancements. The results depicted in the figure underscore the importance of a relatively optimal number of PFEM modules. Overloading the model with excessive PFEM modules may introduce unnecessary complexity without substantive performance gains.

\end{document}